# A Conceptual Model for AI Adoption in Financial Decision-Making: Addressing the Unique Challenges of Small and Medium-Sized Enterprises[1]

Manh-Chiên Vu[1,2][0000-0001-9401-905X], Thang Le Dinh[3][0000-0002-5324-2746], Tran Duc Le[3][0000-0003-3735-0314-2746], Thi Lien Huong Nguyen[2][0000-0002-6512-5053]

[1] University of Quebec at Outaouais, Quebec, Canada
[2] Thuongmai University, Hanoi, Vietnam
[3] University of Quebec at Trois-Rivières, Quebec, Canada

**Abstract.** The adoption of artificial intelligence (AI) offers transformative potential for small and medium-sized enterprises (SMEs), particularly in enhancing financial decision-making processes. However, SMEs often face significant barriers to implementing AI technologies, including limited resources, technical expertise, and data management capabilities. This paper presents a conceptual model for the adoption of AI in financial decision-making for SMEs. The proposed model addresses key challenges faced by SMEs, including limited resources, technical expertise, and data management capabilities. The model is structured into layers: data sources, data processing and integration, AI model deployment, decision support and automation, and validation and risk management. By implementing AI incrementally, SMEs can optimize financial forecasting, budgeting, investment strategies, and risk management. This paper highlights the importance of data quality and continuous model validation, providing a practical roadmap for SMEs to integrate AI into their financial operations. The study concludes with implications for SMEs adopting AI-driven financial processes and suggests areas for future research in AI applications for SME finance.

**Keywords:** AI Adoption, Financial Decision-Making, SMEs, AI model deployment.

## 1 Introduction

Small and medium-sized enterprises (SMEs) are a critical component of economies worldwide, driving innovation, creating jobs, and fostering economic growth.

---

[1] This is a preprint of the following chapter: Manh Chien Vu, Thang Le Dinh, Manh Chien Vu, Tran Duc Le, Thi Lien Huong Nguyen, A Conceptual Model for AI Adoption in Financial Decision-Making: Addressing the Unique Challenges of Small and Medium-Sized Enterprises, published in Artificial Intelligence and Machine Learning for Econometrics: Applications and Regulation (and Related Topics), edited by Nguyen Ngoc Thach, Nguyen Duc Trung, Doan Thanh Ha, and Vladik Kreinovich, 2025, Studies in Systems, Decision, and Control reproduced with permission of Springer Nature

However, SMEs often face challenges in managing their financial processes due to resource constraints, limited access to expertise, and the increasing complexity of the business environment [1]. Financial decision-making processes, such as financial forecasting, budgeting, investment strategies, and risk management, are essential for the survival and growth of SMEs [2, 3], yet these processes are frequently inefficient and prone to errors when conducted manually or with outdated methods [4].

The rapid advancement of artificial intelligence (AI) technologies presents an opportunity for SMEs to optimize their financial decision-making processes [5]. AI can improve accuracy, speed, and efficiency by automating repetitive tasks, analyzing large volumes of data, and providing actionable insights [6]. For example, machine learning algorithms can predict future financial trends, while AI-driven budgeting tools can help optimize resource allocation. These capabilities are particularly valuable for SMEs, which often operate with tight budgets and limited access to sophisticated financial tools.

AI adoption in financial decision-making for SMEs has gained traction in recent years, with various case studies illustrating its benefits [7]. For instance, several SMEs have successfully implemented AI-driven financial forecasting models that leverage historical data to predict sales and cash flows with greater accuracy than traditional methods [8]. AI-powered investment tools have also enabled SMEs to make more informed decisions by analyzing market trends and evaluating risks in real-time. However, despite these successes, the adoption of AI in SME finance remains limited, primarily due to barriers such as high implementation costs, technical complexity, and a lack of tailored solutions for smaller businesses.

The growing body of research on AI in finance tends to focus on large enterprises with abundant resources, leaving a gap in understanding how SMEs can effectively harness AI technologies. This gap presents an opportunity to explore and propose a conceptual model that addresses the unique challenges faced by SMEs in adopting AI for financial decision-making.

Despite the potential benefits, many SMEs struggle to implement AI in their financial processes due to several factors. Key challenges include limited financial resources, a lack of in-house technical expertise, and the perceived complexity of AI technologies. Additionally, existing AI solutions are often designed for larger enterprises and may not be scalable or cost-effective for smaller businesses. These issues create a barrier to entry, preventing many SMEs from realizing the full potential of AI in financial decision-making.

The need for a conceptual model tailored to the specific needs and constraints of SMEs is clear. Such a model should provide a practical roadmap for AI adoption in financial decision-making, considering the limitations of SMEs while leveraging the strengths of AI technologies. This paper aims to develop and propose a conceptual model that addresses these challenges, offering a structured approach to AI adoption that is accessible, scalable, and effective for SMEs. The proposed model stands out by emphasizing practicality and affordability. It provides specific recommendations on how SMEs can start with low-cost, easy-to-implement tools and gradually move toward more advanced AI technologies as their capabilities expand. By focusing on data collection, processing, AI model deployment, decision support, and continuous validation, the model ensures that SMEs can make informed financial decisions with confidence. Moreover, the model highlights the importance of collaboration between AI systems and human decision-makers, fostering a continuous feedback loop that improves the effectiveness of AI applications over time.

The paper is structured as follows. The **Literature Review** section provides an overview of financial decision-making processes in SMEs and explores how AI can enhance these processes. It also discusses the limitations SMEs face in adopting AI for financial decision-making. The **Methodology** section outlines the conceptual modeling approach used to develop the proposed model, detailing the process and layers, such as data sources, model deployment, validation techniques, and risk and sensitivity analysis. The **Proposed Conceptual Model** will be introduced, with a detailed explanation of each layer and the specific tools and techniques that SMEs can use. The methodology behind the model will be outlined, emphasizing its scalability and adaptability to various SME contexts. The **Discussion and Implication** section evaluates the proposed model, comparing it with traditional methods and discussing its practical implications for SMEs. Finally, the **Conclusion** summarizes the key findings, highlights the practical implications of AI adoption for SMEs, and suggests areas for future research.

## 2       Literature review

Artificial Intelligence (AI) has been identified as a critical enabler of more sophisticated financial decision-making processes, particularly in areas such as financial forecasting, budgeting optimization, investment strategies, and risk management [9]. AI's ability to process large volumes of data, identify patterns, and generate actionable insights can significantly improve the accuracy and speed of financial decisions. AI-driven solutions also have the potential to automate routine tasks, allowing finance teams to focus on more strategic functions [10]. While the overall benefits of AI in financial decision-making are widely recognized, it is the specific models, techniques, and methods adopted by enterprises that are instrumental in realizing these benefits.

AI adoption in financial decision-making across enterprises has primarily been driven by the implementation of various machine learning models, predictive analytics, natural language processing (NLP), robotic process automation (RPA), and deep learning techniques. These technologies have reshaped traditional financial processes by introducing automation, real-time analytics, and advanced data processing capabilities.

**Machine learning (ML)** models have been central to AI adoption in financial decision-making. Supervised learning models, such as linear regression and decision trees, have been widely utilized for tasks like financial forecasting and risk assessment. For instance, decision trees are employed to predict future revenue or identify potential risks by analyzing historical financial data. Enterprises have adopted these models to enhance the accuracy of financial forecasts, allowing for more informed decision-making [11, 12]. Unsupervised learning models, such as clustering algorithms, are also used to detect anomalies, particularly in fraud detection and risk management. These models can process vast amounts of transaction data to identify patterns that deviate from the norm, flagging potential fraudulent activities for further investigation.

In the realm of investment strategies, **reinforcement learning (RL)** has emerged as a powerful tool. Enterprises are increasingly utilizing RL algorithms to develop dynamic investment strategies that adapt to changing market conditions. RL models continuously learn from their interactions with the financial environment, adjusting their strategies based on feedback from the market. This method is particularly valuable in portfolio management, where AI-driven strategies can optimize investment decisions in real-time by analyzing market data and identifying opportunities for returns while mitigating risks [13].

**Predictive analytics** is another key AI technique that has been extensively adopted for financial decision-making, particularly in forecasting and budgeting. Time series analysis models, such as ARIMA (AutoRegressive Integrated Moving Average) and Prophet, have been employed by enterprises to forecast financial metrics such as revenue, sales, and expenses. These models analyze historical financial data to predict future trends, enabling businesses to make data-driven decisions. Additionally, regression models, both linear and nonlinear, are widely used to predict how changes in external variables, such as market conditions or customer demand, will impact financial performance. This allows enterprises to adjust their budgeting and financial planning strategies accordingly [14].

Advanced **ensemble methods** have also gained traction in financial decision-making. These methods combine the outputs of multiple models to improve prediction accuracy. For example, techniques like Random Forest and Gradient Boosting aggregate the results of different machine learning models, thereby reducing the risk of overfitting and providing more robust financial forecasts. Enterprises have adopted these ensemble methods to enhance the precision of their financial predictions, which is crucial for making strategic decisions [15, 16].

In addition to machine learning, **Natural Language Processing (NLP)** has been adopted in financial decision-making, particularly in analyzing unstructured data such as financial reports, news articles, and legal documents. NLP techniques, such as text mining and sentiment analysis, allow enterprises to gauge market sentiment by analyzing public statements, earnings calls, and media coverage. Sentiment analysis, for example, can influence investment strategies by identifying market trends that could impact stock prices or economic conditions. Moreover, NLP-driven automation tools have been employed to streamline financial reporting, compliance, and contract analysis. By processing large volumes of textual data, these tools significantly reduce the time and effort required by human analysts, allowing finance teams to focus on higher-level strategic activities [17].

**Robotic Process Automation (RPA)** has further revolutionized financial decision-making by automating repetitive and rule-based tasks. Enterprises have widely adopted RPA to streamline processes such as transaction processing, invoice management, payroll, and account reconciliation. For example, in invoice processing, RPA bots can automatically extract data from invoices, validate the information, and process payments, significantly reducing manual effort and minimizing errors. Similarly, RPA has been used to automate payroll processes, ensuring accurate and timely payments while complying with tax regulations. By automating these routine tasks, RPA enables finance teams to allocate more resources to strategic decision-making and problem-solving [18].

**Deep learning**, a subset of machine learning, has also found applications in more complex financial decision-making tasks. Deep learning models, such as convolutional neural networks (CNNs) and recurrent neural networks (RNNs), are capable of processing vast amounts of structured and unstructured data, identifying patterns that traditional models might overlook. For instance, RNNs, particularly Long Short-Term Memory (LSTM) networks, have been employed in financial forecasting tasks where sequential data, such as stock prices, need to be analyzed over time. These models excel at capturing temporal dependencies and are therefore well-suited for predicting future financial trends [19]. In fraud detection, deep learning models such as CNNs are used to analyze complex transaction data and identify patterns indicative of fraudulent activity. These models can process large datasets in real-time, making them valuable for detecting and preventing fraud in financial transactions [20].

The adoption of these AI models and techniques in enterprises has led to significant improvements in financial decision-making, from increasing the accuracy of forecasts to automating routine financial tasks. However, while large enterprises have successfully integrated these advanced AI tools, smaller enterprises, such as SMEs, face unique challenges that hinder the adoption of AI.

Despite the demonstrated benefits of AI adoption in financial decision-making, SMEs encounter several barriers that limit their ability to fully leverage AI technologies. A significant challenge is the **resource constraint** faced by SMEs [21]. Unlike larger enterprises with substantial financial resources, SMEs operate with tight budgets that may not accommodate the costs associated with implementing AI systems. The expenses involved in purchasing AI software, maintaining the necessary infrastructure, and training employees to use AI tools can be prohibitive. This financial limitation often forces SMEs to prioritize other business needs over investing in AI, even when the potential return on investment is clear [22].

In addition to financial constraints, SMEs often lack the **technical expertise** required to implement and manage AI systems effectively. AI technologies, particularly machine learning and deep learning models, require specialized knowledge in data science, algorithms, and software development. SMEs may not have the in-house expertise needed to develop, deploy, and maintain AI models, and recruiting or upskilling employees can be both costly and time-consuming. This lack of expertise can result in poorly implemented AI solutions that fail to deliver the expected benefits [22].

Another challenge is the **complexity of AI technologies** and the difficulty of integrating them with existing systems. AI solutions can be complex to deploy and may require significant modifications to an organization's IT infrastructure. SMEs, which often have less sophisticated IT systems than larger enterprises, may struggle to integrate AI technologies seamlessly. This complexity can create additional barriers to adoption, as SMEs may not have the internal resources or technical support needed to implement these technologies effectively [22].

The issue of **data quality and availability** also poses a significant challenge for SMEs adopting AI. AI models rely on large datasets to function effectively, but SMEs may have limited access to the volume and quality of data needed to train these models.

Inadequate data collection processes, incomplete datasets, and poor data quality can undermine the effectiveness of AI models, leading to inaccurate or unreliable outputs. Without access to high-quality data, SMEs may find that AI models do not deliver the expected improvements in financial decision-making [23].

## 3  Methodology

A conceptual modeling approach provides a structured framework for understanding and representing complex systems, making it particularly useful in fields such as finance, where decision-making involves multiple interconnected processes [24]. Conceptual models are valuable not only as theoretical constructs but also as practical guides for implementation, offering a roadmap that organizations can follow as they adopt new technologies. In the context of AI adoption for financial decision-making in SMEs, a conceptual model helps to visualize and organize the various components of AI integration - from data collection and processing to model deployment and decision support - into a coherent system. This structured approach is especially critical for SMEs, which often face resource constraints, limited technical expertise, and the complexity of integrating AI with existing systems. By breaking down the AI adoption process into manageable layers, a conceptual model enables SMEs to approach AI integration incrementally, ensuring that each phase is carefully planned and executed. This methodology facilitates a more accessible and scalable implementation, addressing the unique challenges that SMEs face in adopting AI-driven financial solutions.

In developing the proposed conceptual model for AI adoption in SME financial decision-making, a multi-step process was followed, grounded in both theoretical insights and practical considerations. The model aims to provide a simplified yet comprehensive framework for SMEs, enabling them to gradually implement AI technologies in key financial processes. The first step involved identifying the core financial decision-making processes in SMEs, such as financial forecasting, budgeting, investment strategy formulation, and risk management. Improving the efficiency and accuracy of these processes through AI-driven solutions is central to the model's objective. The study then mapped relevant AI technologies to these financial processes, carefully selecting models, methods, and techniques that are both practical and scalable for SMEs.

The conceptual model is designed as a layered framework, reflecting the various stages of AI adoption and integration. Each layer addresses a specific aspect of the AI implementation process, from data collection and processing to model deployment, validation, and decision support. This layered structure allows SMEs to adopt the model incrementally, scaling their AI capabilities as their needs evolve. Through continuous validation and feedback loops, the model also integrates risk and sensitivity analysis, helping SMEs navigate uncertainties and improve the accuracy of their financial decisions over time.

## 4      Proposed model

The conceptual model for AI adoption in financial decision-making for small and medium-sized enterprises (SMEs) is designed to be both practical and adaptable. It recognizes the unique challenges faced by SMEs - such as limited financial resources, a lack of technical expertise, and the complexity of integrating AI systems. To address these challenges, the model offers a step-by-step approach that allows SMEs to incrementally implement AI technologies. This layered model focuses on specific, actionable steps, tools, and technologies, helping SMEs transition from basic data management to advanced AI-driven financial decision-making.

The model consists of several interconnected layers, each focusing on a critical aspect of the AI adoption process. These layers are designed to be modular, allowing SMEs to implement them incrementally based on their resources and capabilities. The layers include the Data Sources Layer, Data Processing and Integration Layer, AI Model Deployment Layer, Decision Support and Automation Layer, and Validation and Risk Management Layer. Together, these layers form a cohesive framework for integrating AI into financial decision-making processes.

The foundation of the model lies in the **Data Sources Layer**, which highlights the importance of collecting and integrating relevant financial data. SMEs can begin with simple data collection methods using basic tools such as Google Sheets, Excel, or widely used accounting software like QuickBooks. These tools enable SMEs to gather and organize internal data, including financial records, sales information, and customer data. In addition to internal data, external sources such as industry reports, market trends, and publicly available datasets can be utilized. As SMEs grow and their data needs increase, they can scale up to more sophisticated data management platforms, such as cloud-based databases like Google Cloud SQL or Amazon RDS. These platforms allow SMEs to manage larger datasets and ensure that their data is consistently available and properly integrated [25]. The focus in this foundational layer is on maintaining data quality and completeness, as AI models rely heavily on accurate input data for reliable outcomes.

Once the data has been collected, it moves to the **Data Processing and Integration Layer**. In this stage, raw data must be cleaned, transformed, and integrated to prepare it for analysis by AI models. SMEs can start with accessible tools like Excel or Python's Pandas library for basic data cleaning and normalization. These tools help eliminate errors, manage missing data, and standardize formats across datasets. As the business scales, more robust data processing platforms, such as Apache NiFi or Microsoft Power BI, can be integrated to automate workflows and manage data processing more efficiently. Cloud storage solutions like Dropbox Business or Google Cloud Storage provide scalable options for secure data management, which is critical as SMEs accumulate more data over time. This layer ensures that the data fed into AI models is of high quality, well-structured, and readily accessible [26].

The core of the conceptual model is the **AI Model Deployment Layer**, where AI technologies are applied to enhance financial decision-making. Pre-built AI services from platforms like Microsoft Azure AI, Google AI, or IBM Watson are particularly valuable for SMEs, as they reduce the complexity and cost of implementation [27].

These services offer AI models that can be tailored to various financial needs, including forecasting, budgeting, and risk analysis. For instance, Microsoft's AI Builder allows users to create custom AI models directly within Power Apps, making advanced AI technologies accessible even to those without extensive technical expertise. Continuous model training and deployment are key elements in this layer, ensuring that AI systems remain adaptive to evolving financial conditions. As SMEs expand their AI capabilities, they can transition to more sophisticated platforms like Amazon SageMaker or DataRobot, which offer machine learning capabilities tailored to specific financial tasks. Monitoring and maintaining the performance of AI models is essential to ensure that the insights they generate remain accurate and relevant.

Once the AI models are deployed, the focus shifts to the **Decision Support and Automation Layer**, where AI-generated insights are integrated into the daily financial operations of SMEs. AI models provide real-time recommendations for financial forecasting, cash flow management, budgeting optimization, and investment strategies. SMEs can start with accessible AI-driven financial tools like Fathom or Float, which integrate with existing accounting software to provide insights. In addition to decision support, this layer emphasizes automation. Robotic Process Automation (RPA) tools like UiPath can automate routine financial tasks, such as transaction processing, account reconciliation, and payroll management. By automating these repetitive tasks, SMEs can focus their resources on more strategic financial decision-making [28].

The **Validation and Risk Management Layer** addresses the need to ensure the reliability and accuracy of AI-driven financial decisions. Validation techniques, such as cross-validation, are used to assess the performance of AI models. SMEs can utilize the built-in validation features available in platforms like Google AI and Microsoft AI Builder to monitor and adjust their models [29]. For risk management, tools like Adaptive Insights or Planful provide AI-driven financial planning that helps SMEs identify potential risks and assess how different variables might impact their financial outcomes. As SMEs mature in their AI adoption, they can implement more advanced risk management platforms such as RiskWatch or Palisade's RISK software, which offer comprehensive risk analysis capabilities. This layer ensures that AI-driven insights are trustworthy and that potential risks are effectively managed.

The final layer of the model, the **Collaboration and Feedback Layer**, focuses on fostering collaboration between human decision-makers and AI systems. AI adoption is an ongoing process that requires regular updates, feedback, and refinements. SMEs should prioritize collaboration by integrating tools like Microsoft Teams or Slack with AI dashboards. These tools facilitate seamless communication and feedback loops, ensuring that AI insights are continuously refined based on real-time inputs from finance teams. Shared dashboards, such as those provided by Tableau or Zoho Analytics, allow teams to visualize AI-generated insights and collaborate on decision-making in a more interactive and dynamic way. Continuous feedback ensures that AI models improve over time and remain aligned with the business's financial goals [30].

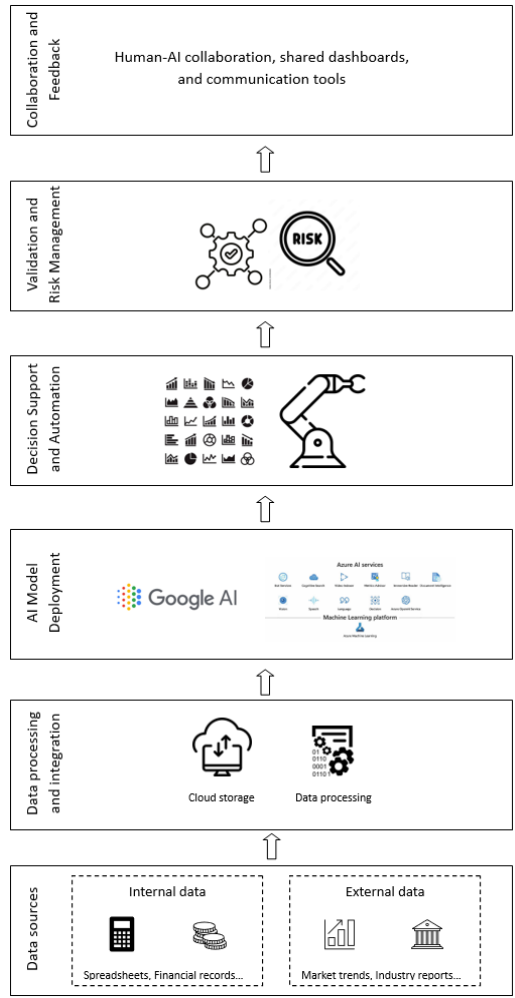

**Fig. 1.** Conceptual Model for AI Adoption in Financial Decision-Making for SMEs

## 5 Discussions and Implications

Traditional financial decision-making methods in SMEs often rely on manual processes, such as spreadsheets and basic forecasting techniques. These methods are typically reactive and based on historical data, lacking the predictive capabilities required to anticipate future trends effectively. AI-driven financial decision-making models, on the other hand, enable a more proactive approach by utilizing advanced algorithms to analyze large volumes of data in real time. By identifying patterns and trends that traditional methods may overlook, AI can significantly enhance the accuracy and timeliness of financial decisions [31].

One of the primary advantages of the proposed model is its ability to automate routine financial tasks, such as budgeting, forecasting, and transaction processing. Automation not only reduces the time and effort required for these tasks but also improves consistency and minimizes the risk of human error. Furthermore, AI models can continuously learn and improve as they process more data, becoming more efficient and accurate over time. In contrast, traditional methods require ongoing manual updates, which can be time-consuming and error-prone [32].

However, the adoption of AI technologies also presents challenges for SMEs, particularly in terms of resource allocation and technical expertise. The conceptual model addresses these challenges by offering a step-by-step approach that allows SMEs to implement AI incrementally, reducing the risk of failure and ensuring that each phase of AI integration is fully integrated before moving on to the next. By starting with affordable, easy-to-use tools and scaling up as their capabilities grow, SMEs can benefit from AI-driven decision-making without overextending their resources.

The proposed model has several strengths that make it particularly well-suited for SMEs. First, its modular design allows for incremental implementation, enabling SMEs to adopt AI technologies gradually as their financial and technical capabilities expand. This approach minimizes upfront costs and reduces the risks associated with large-scale AI projects. Additionally, the model emphasizes the importance of selecting tools and technologies that are affordable and easy to use, ensuring that SMEs can access AI solutions even with limited budgets [33].

Another strength of the model is its focus on data quality and integration. By prioritizing data collection, processing, and management, the model ensures that AI-driven decisions are based on accurate and reliable information. This focus on data quality is critical for the success of AI models, as poor-quality data can lead to inaccurate predictions and flawed decision-making [34]. The use of cloud-based data management platforms further enhances the model's scalability and cost-effectiveness, making it accessible to SMEs that need flexible and affordable solutions.

Despite these strengths, the model also has limitations. One of the primary challenges is the need for continuous monitoring and validation of AI models. While AI-driven decision-making can significantly improve financial processes, it requires ongoing oversight to ensure that models remain accurate and aligned with business goals. SMEs may find this aspect of AI adoption challenging, particularly if they lack the necessary technical expertise or resources to continuously monitor and adjust AI models [31]. Additionally, although the model emphasizes ease of use, some SMEs may still face resistance to adopting AI-driven solutions, especially in organizations where manual decision-making processes are deeply ingrained.

**Practical and Strategic Implications for SMEs**
The adoption of AI-driven financial decision-making models can have profound practical and strategic implications for SMEs. On a practical level, AI technologies can improve the efficiency and accuracy of financial management by automating routine tasks and providing real-time insights. This allows finance teams to make more informed decisions, optimize resource allocation, and better manage cash flow. In turn, this can lead to more effective budgeting, reduced financial risk, and improved overall financial performance [32].

Strategically, AI adoption can give SMEs a competitive advantage by enabling them to respond more quickly and effectively to market changes. AI-driven insights can help SMEs identify growth opportunities, optimize their operations, and adapt to evolving market conditions. This agility is particularly important in today's fast-paced business environment, where SMEs must be able to pivot quickly to remain competitive. Additionally, AI technologies can help SMEs better manage financial risks, providing them with the tools they need to navigate complex financial landscapes and make more strategic, data-driven decisions [30].

However, to fully realize these benefits, SMEs must carefully manage the challenges associated with AI adoption. This includes investing in the necessary infrastructure, training staff to work with AI technologies, and fostering a culture of innovation and adaptability. SMEs that successfully integrate AI into their financial decision-making processes will be well-positioned to thrive in an increasingly data-driven business environment. Furthermore, the flexibility of the proposed model allows SMEs to tailor AI adoption to their specific needs, ensuring that the benefits of AI are maximized without overwhelming the organization.

**Future Research Directions**

While this paper proposes a practical conceptual model for AI adoption in SME financial decision-making, further research is needed to explore the long-term impact of AI integration on SME performance. Future studies should focus on conducting longitudinal analyses to assess the effectiveness of AI-driven financial models over time. Additionally, there is a need for research that develops AI solutions specifically tailored to the needs of SMEs, considering their resource constraints and unique business challenges [31].

Another important area for future research is the exploration of ethical considerations related to AI adoption in finance. As AI technologies become more prevalent in financial decision-making, ensuring transparency, fairness, and compliance with regulatory requirements will be critical. Research into the ethical implications of AI in finance can help guide the development of responsible AI solutions that benefit both SMEs and the broader economy [34].

Finally, further research should explore best practices for AI implementation in SMEs. Identifying effective strategies for overcoming barriers to AI adoption, such as cost, complexity, and resistance to change, can help bridge the gap between AI's potential and its real-world application in SMEs. Providing SMEs with practical guidance on how to implement AI successfully will be key to unlocking the full potential of AI in financial decision-making.

## 6   Conclusion

The proposed conceptual model for AI adoption in financial decision-making provides small and medium-sized enterprises (SMEs) with a practical and scalable framework to integrate AI technologies into their operations. Recognizing the unique challenges faced by SMEs, including limited resources, technical expertise, and the complexity of AI integration, this model offers a step-by-step approach that is both adaptable and actionable. By emphasizing affordability, incremental implementation, and scalability, the model ensures that SMEs can begin with basic AI tools and gradually advance to more sophisticated systems as their capabilities expand.

One of the key strengths of the model is its focus on guiding SMEs through the AI adoption process in a structured manner. Starting with essential layers such as data collection and processing, the model progressively introduces more advanced AI model deployment and decision support tools. This layered approach helps SMEs manage the complexities of AI adoption without overwhelming their resources. By using affordable and accessible technologies, the model makes AI-driven financial decision-making attainable for SMEs that may not have the capacity to invest in large-scale AI solutions.

Additionally, the model highlights the importance of continuous validation, risk management, and collaboration between AI systems and human decision-makers. These elements ensure that AI models remain accurate and relevant to the business's financial goals, while also fostering a culture of ongoing improvement. The focus on practical implementation, coupled with the flexibility to tailor AI adoption to specific SME needs, positions this model as a valuable tool for enhancing financial management in smaller enterprises.

However, while the model provides a solid foundation for AI adoption, further research is needed to assess its long-term impact on SME performance. Future studies should explore how AI-driven financial models perform over time and investigate the development of AI solutions specifically designed for SMEs. Additionally, as AI technologies become more embedded in financial decision-making, ethical considerations - such as transparency, fairness, and regulatory compliance - will need to be addressed to ensure responsible AI use in finance.

Ultimately, the proposed conceptual model offers SMEs a feasible and impactful roadmap for leveraging AI to enhance their financial decision-making. By starting small and scaling up as their capabilities grow, SMEs can gradually integrate AI into their operations, improving forecasting accuracy, budgeting efficiency, and strategic financial management. This model bridges the gap between the potential of AI and its practical application in SMEs, empowering smaller enterprises to compete more effectively in an increasingly data-driven business environment.

**Funding:** This research is funded by Thuongmai University, Hanoi, Vietnam